\documentclass[10pt,twocolumn,letterpaper]{article}

\usepackage{wacv}              

\usepackage{graphicx}
\usepackage{amsmath}
\usepackage{amssymb}
\usepackage{booktabs}
\usepackage{hhline}

\usepackage[pagebackref,breaklinks,colorlinks]{hyperref}
\usepackage{multirow, adjustbox, makecell}
\usepackage{algorithm}
\usepackage{amssymb}
\usepackage{relsize}
\usepackage{algpseudocode}
\usepackage{boldline}
\usepackage{subcaption}
\usepackage[dvipsnames, table]{xcolor}
\usepackage[hang,flushmargin]{footmisc}
\usepackage{multibib}
\usepackage{enumitem}
\makeatletter
\robustify\@latex@warning@no@line
\makeatother
\usepackage{authblk}

\usepackage[capitalize]{cleveref}
\crefname{section}{Sec.}{Secs.}
\Crefname{section}{Section}{Sections}
\Crefname{table}{Table}{Tables}
\crefname{table}{Tab.}{Tabs.}

\def\wacvPaperID{122} 
\def\confName{WACV}
\def\confYear{2025}

\begin{document}

\title{GEXIA: Granularity Expansion and Iterative Approximation \\for Scalable Multi-grained Video-language Learning} 



\makeatletter \renewcommand\AB@affilsepx{\quad \protect\Affilfont} \makeatother

\author[1,2\thanks{Work was done as an intern at Amazon.}]{Yicheng Wang}
\author[2]{Zhikang Zhang}
\author[2]{Jue Wang}
\author[2]{David Fan}
\author[2]{\\Zhenlin Xu}
\author[2]{Linda Liu}
\author[2]{Xiang Hao}
\author[2]{Vimal Bhat}
\author[2]{Xinyu Li}

\affil[1]{Texas A\&M University}
\affil[2]{Amazon}


\maketitle
 \begin{abstract}
In various video-language learning tasks, the challenge of achieving cross-modality alignment with multi-grained data persists. We propose a method to tackle this challenge from two crucial perspectives: data and modeling. Given the absence of a multi-grained video-text pretraining dataset, we introduce a Granularity EXpansion (GEX) method with Integration and Compression operations to expand the granularity of a single-grained dataset. To better model multi-grained data, we introduce an Iterative Approximation Module (IAM), which embeds multi-grained videos and texts into a unified, low-dimensional semantic space while preserving essential information for cross-modal alignment. Furthermore, GEXIA is highly scalable with no restrictions on the number of video-text granularities for alignment. We evaluate our work on three categories of video tasks across seven benchmark datasets, showcasing state-of-the-art or comparable performance. Remarkably, our model excels in tasks involving long-form video understanding, even though the pretraining dataset only contains short video clips.
\end{abstract} 
 \section{Introduction}
\label{sec:intro}

Video-language representation learning is crucial for a wide range of video tasks, such as cross-modality retrieval~\cite{activitynet, msrvtt, lsmdc}, video classification~\cite{lvu, coin, charades}, and video question answering~\cite{hero}. With the advent of CLIP-like models~\cite{clip,scaleupclip,Alayrac2022FlamingoAV, clip_multi}, video-language learning has entered the "big data" era, where video-text foundation models~\cite{lfvila, clip4clip, clipbert, frozen, taco, hiervl} are first pretrained on large-scale datasets consisting of diverse video-text pairs aimed at cross-modal representation alignment and then finetuned on small-scale datasets tailored to specific downstream tasks. Following this paradigm, recent research primarily focuses on two main directions: data and modeling. 

On the data front, researchers have introduced larger-scale pretraining datasets~\cite{howto100m, internvid, hdvila} with higher-quality video-text pairs using scalable data collection methods. On the modeling side, most approaches~\cite{clip4clip, clipbert, lfvila, taco, hiervl, hitea, frozen} focus on single-grained or hierarchical video-text representation alignments. Nevertheless, a frequently overlooked fact should be revisited: video-text data are inherently multi-grained, i.e., videos and texts in the real world naturally have variable durations and text sequence lengths, while the same video can also be paired with texts in different sequence lengths, offering varying levels of detail, and vice versa. This inherent multi-granularity presents a major challenge for modeling video-text alignment. Previous methods~\cite{clip4clip, clipbert, taco, hiervl, lfvila, hitea,fan2025text,lee2024video} often fall short of fundamentally addressing the challenge.


\begin{figure*}[!h]
    \centering
    \includegraphics[width=0.9\textwidth]{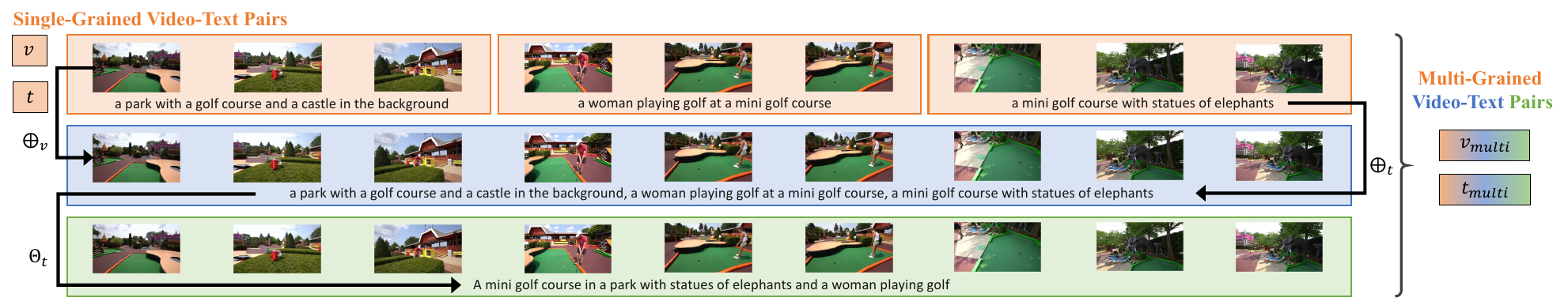}
    \vspace{-2mm}
    \caption{An overview of the Granularity EXpansion (GEX) pipeline, which expands a single-grained dataset into a multi-grained dataset with video and text integration $\oplus_v$ and $\oplus_t$ and text compression $\Theta_t$ operations.}
    \label{fig1}
    \vspace{-3mm}
\end{figure*}


In this work, we propose \textbf{GEXIA}: \textbf{G}ranularity \textbf{EX}pansion and \textbf{I}terative \textbf{A}pproximation to address this challenge from both data and modeling perspectives. Firstly, existing video-text pre-training datasets~\cite{howto100m, frozen, hdvila, internvid} follow data collection schemas similar to image-text datasets. As a result, these datasets either provide only seconds-long videos with sentence-long texts~\cite{internvid,frozen} or minutes-long videos with audio transcriptions as texts~\cite{hdvila,howto100m}. Constrained by scalability, such datasets are essentially single-grained and cannot reflect the multi-granularity of real-world video-text data, which hinders model performance during pretraining. To this end, we introduce a scalable video-text Granularity EXpansion (GEX) pipeline that generates multi-grained video-text datasets from single-grained ones. Specifically, we propose Integration and Compression operations on videos and texts separately to expand two additional granularities by using only existing samples in the dataset. These operations can be applied sequentially to achieve the desired number of granularities, as shown in Figure~\ref{fig1}.




Secondly, most recent works~\cite{clip4clip, frozen, taco, lfvila, clipbert} treat input video-text pairs as single-grained regardless of differences in video durations and text lengths, which limits their performance in tasks that involve multi-grained data. Other studies~\cite{hitea, hiervl, xclip,videorecap} address the issue using hierarchical designs to handle different granularities of data at varying structural levels. However, such approaches face significant limitations in scalability to new granularities or are tailored for specific downstream tasks only. A comprehensive solution for multi-grained video-text alignment across diverse downstream tasks with high scalability has not been thoroughly explored. Aiming at the modeling challenge, we propose an Iterative Approximation Module (IAM) as a general building block for multi-grained video-text alignment presented in Figure~\ref{fig2}. We formulate the multi-grained alignment as a three-stage process: dense feature extraction, iterative approximation, and cross-model alignment. First, dense features of the videos and texts are extracted and aggregated separately with an existing image and text encoder. Subsequently, IAMs iteratively and adaptively approximate these dense features into fixed-size low-dimensional embeddings, considering the granularity of inputs. As such, uniform representations of multi-grained videos and texts are embedded in the same low-dimensional space to enable the final cross-modal alignment through conventional contrastive learning.


We take an existing single-grained pretraining dataset to showcase the effectiveness of GEX and conduct extensive evaluations of IAMs across cross-modal retrieval, video classification, and video-question-answering tasks on \textbf{seven} widely used datasets of various granularities. Our model, enhanced by the multi-grained pretraining data, consistently achieves state-of-the-art or comparable performance on all tasks compared to baseline models. Notably, our model excels in scenarios requiring long-term visual-text alignment and a deep understanding of long-form videos, highlighting the effectiveness of our approach. In summary, our contributions are as follows.


\begin{enumerate}

\vspace{-1mm}
\item We propose Granularity EXpansion (GEX), a scalable method to automatically expand single-grained video-text datasets into multi-grained ones without additional data collection, addressing the multi-grained pretraining data bottleneck;
\vspace{-2mm}
\item We introduce the Iterative Approximation Module (IAM), which embeds multi-grained data into a unified, low-dimensional space, enabling efficient and scalable cross-modal alignment;
\vspace{-2mm}
\item Our approach achieves state-of-the-art or comparable results across multiple benchmarks. We also provide detailed ablations and insights.
\vspace{-1mm}
\end{enumerate}

 \section{Related work}
\label{sec:related}
\subsection{Single-grained Video-Language Learning}
Early works~\cite{clipbert,taco,frozen,clip4clip,perceivervl} have treated input video and text pairs as in single granularity regardless of the video duration and text length. Due to the overlook of inherent multi-granularity, such approaches are not only limited to tasks that involve multi-grained video-text data but also sacrifice performances on general video-text alignments. More specifically, CLIPBert~\cite{clipbert} proposes to apply sparse sampling on videos in a universal manner, which leads to poor performance on long-form video understanding tasks. TACo \cite{taco} relies on heuristics that nouns and verbs are more likely to be aligned with visual contents, which sacrifices the performance on long, complicated video scenes and texts. Frozen~\cite{frozen} views images as static video clips and pretrain on images and videos simultaneously, whose performance is limited on long videos with large temporal dynamics. Clip4clip~\cite{clip4clip} empirically studies the approaches of aggregating image-level CLIP features for video retrieval tasks through mean-pooling, LSTM, or single transformer block, which is largely limited to video retrieval tasks and only achieves coarse alignments between videos and text. LF-VILA~\cite{lfvila} focuses on single granularity video-text alignment for long-form video understanding. Despite the fact that LF-VILA is limited to long-form videos only, our approach uses only 1/6 of video-text pairs for pretraining while achieving comparable or better performance than LF-VILA on multiple long-form video understanding tasks as shown in Section~\ref{sec:experiment}.
\subsection{Multi-grained Video-Language Learning}
Observing the inherent multi-granularity in video-text alignment, 
A few recent works~\cite{xclip,hiervl,hitea,videorecap} address the multi-grained video-text alignment problem either with a hierarchical design, which limits scalability or target specific downstream tasks only, which limits generalizability. X-CLIP~\cite{xclip} targets video retrieval tasks only and proposes a multi-grained contrastive learning loss and an attention-over-similarity-matrix-module to aggregate multi-grained features, which is limited to retrieval tasks only and unable to extend to video-text granularities beyond seconds-long videos and sentence-long texts. HierVL~\cite{hiervl} builds hierarchical video-text embeddings for multi-level video-text alignments and studies two aggregation designs to acquire features of long context windows. This method heavily depends on a high-quality multi-grained video-text pretraining dataset, which is often hard to acquire from the public domain, and can not be well extended to new granularities (HierVL is built on Ego4D, whose videos are specifically collected from non-public domain and texts are written by human narrators). HiTeA~\cite{hitea} proposes to achieve multi-grained video-text alignments by designing two new pre-training tasks, namely cross-modal moment exploration and multi-modal temporal relation exploration. Such tasks extensively extract temporal information from input video sequences, while leading to limitations on long-form videos and more video-text granularities. VideoRecap~\cite{videorecap} specifically targets the video captioning problem and also employs a hierarchical design to process multi-grained data. 
\section{Methodology}
\label{sec:all_method}

\subsection{Granularity EXpansion (GEX)}
\label{sec:data}
In real-world application scenarios, videos and texts exhibit various levels of granularity. However, due to the scalability constraints of data collection processes, existing large-scale video-text pretraining datasets \cite{howto100m, frozen, hdvila, internvid} are mostly single-grained. To overcome the challenge of insufficient multi-grained pretraining data, we introduce a simple and scalable Granularity EXpansion (GEX) pipeline to enhance the granularity of an existing single-grained dataset, as illustrated in Figure~\ref{fig1}. The GEX pipeline comprises two main operations: Integration and Compression, which can be further categorized based on the input data type as Video Integration ($\oplus_v$), Text Integration ($\oplus_t$), Video Compression ($\Theta_v$), and Text Compression ($\Theta_t$). 


\noindent\textbf{Integration ($\oplus_v$ and $\oplus_t$):} 
The Integration operation combines multiple input video-text pairs of the same granularity into one data sample to create a new granularity. Specifically, we exploit meta-information to identify existing video-text pairs originating from the same source and concatenate ($\oplus_v$ and $\oplus_t$) them to generate longer video-text pairs. 
When meta-information is unavailable, random samples can also be concatenated in a similar manner, which still enhances the performance, as shown in Appendix A.



\noindent\textbf{Compression ($\Theta_v$ and $\Theta_t$):}
The Compression, in contrast, condenses individual data samples into more compact forms to create a new granularity. Specifically, we utilize text summarization via a large language model (LLM) as $\Theta_t$ to condense $\oplus_t$-applied long texts. The summarized texts are much shorter but remain aligned with the corresponding $\oplus_v$-applied long videos, thereby creating a new granularity. The summarized text is verified to be similar with the original long texts, as demonstrated in Appendix B. We do not perform the video compression $\Theta_v$ in the main study, since $\oplus_v$, $\oplus_t$, and $\Theta_t$ are sufficient to transform a single-grained dataset into a multi-grain dataset for pretraining. Additional exploration of video compression ($\Theta_v$) is included in Appendix F.

GEX is highly scalable and does not rely on human annotation. Both Integration $\oplus$ and Compression $\Theta$ operations have no limits to the input data and can be applied sequentially or recursively to expand the granularities of the given dataset to a desired scale. Using GEX, we can generate a multi-grained video-text pretraining dataset $(v_{multi}, t_{multi})$ from a single-grained dataset $(v, t)$, as shown in Figure~\ref{fig1}.

\begin{figure*}[!t]
    \centering
    \includegraphics[width=0.9\textwidth]{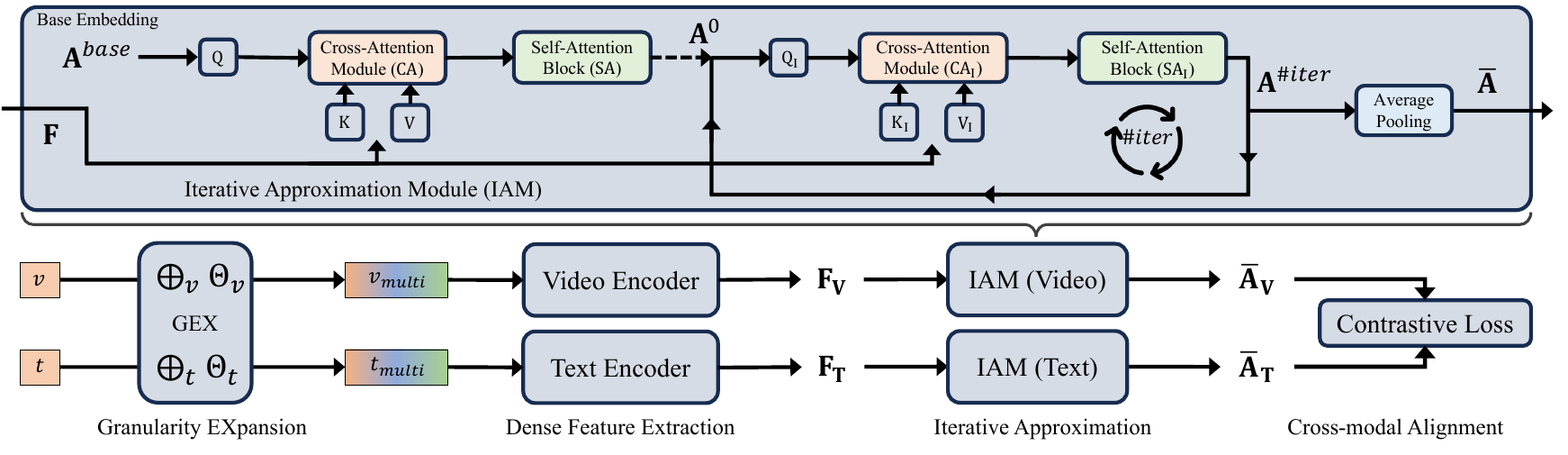}
    \vspace{-3mm}
    \caption{An overview of GEXIA, which consists of Granularity EXpansion (GEX), Dense Feature Extraction, Iterative Approximation Module (IAM), and Cross-modal Alignment. We propose GEX and IAM to address data and modeling challenges, respectively.}
    \vspace{-4mm}
    \label{fig2}
\end{figure*}
\subsection{Multi-grained Video-Text Alignment}
\label{sec:method}

\noindent\textbf{Overview:}
We introduce a three-phase approach for multi-grained video-text alignment, as illustrated in Figure~\ref{fig2}. First, we utilize two distinct single-modality encoders for videos and texts to create high-dimensional representations that encapsulate the complete information of the input data, referred to as dense features. The dimensions of these dense features are proportional to the granularity of the input data, whether determined by the video durations or text sequence lengths. In the second phase, given pairs of dense features with variable dimensions, we aim to approximate these features using fixed-size embeddings derived from a universal low-dimensional space. Apparently, higher-dimensional features with richer semantic information are inherently more challenging to approximate with fixed-size embeddings. We employ cross-attention mechanism as the building block to construct an iterative approximation module, thereby allocating more iterations to higher-dimensional dense features for finer approximation. Finally, given multi-grained videos and texts embedded in the same low-dimensional space, we apply conventional contrastive learning to enforce cross-modal alignment.

\noindent\textbf{Dense Feature Extraction:}
\label{sec:feature_encoder}
For video-text pairs with varied granularity, the first step is to construct feature representations that capture their complete semantic information. Considering the varying lengths of videos and texts, we propose that the dimensionality of these representations should be proportional to the input size, maintaining consistent information density and minimizing information loss. Additionally, since our goal is to create aligned representations across modalities, it is crucial to keep each data stream separate to prevent information leakage. Therefore, we employ pretrained image and text encoders to extract individual features at the frame and text token levels, respectively, and subsequently aggregate these features along the video temporal and text sequential dimensions to form dense features.

Given $d$ video frames $[f_1,\cdots,f_d]$ sampled from a video in $v_{multi}$, where each frame $f_i \in \mathbb{R}^{H\times W\times 3}$, we first employ a ViT-based image encoder to transform each frame into a tensor in $\mathbb{R}^{P\times C_v}$, where $P$ denotes the patch number and $C_v$ denotes the video feature dimension. Then we concatenate all frame-level features into a tensor $\mathbf{F_V}\in \mathbb{R}^{(d\times P) \times C_v}$. Similarly, given the corresponding text sequence $t_{multi}$, we tokenize it using byte pair encoding (BPE) and then truncate or pad it into a size of $m\times S$, where $m$ is the maximum number of tokens that the text encoder can process in a single run and $S$ is the size of one token. Subsequently, we apply the text encoder to transform it into token-level dense feature $\mathbf{F_T}\in \mathbb{R}^{m\times C_t}$, where $C_t$ represents the size of the feature.

\noindent\textbf{Iterative Approximation:}
\label{sec:IAE_module}
Given variable-sized dense features, we develop a novel Iterative Approximation Module (IAM) to approximate the inputs into fixed-sized and low-dimensional embeddings. An IAM takes a dense feature $\mathbf{F} \in \mathbb{R}^{M \times C}$ and a learnable base embedding $\mathbf{A}^{base} \in \mathbb{R}^{N \times D}$ as inputs, where $N$ is the number of learnable base vectors and $D$ is the dimension. In the video IAM, the notation $\mathbf{F}\in \mathbb{R}^{M \times C}$ corresponds to $\mathbf{F_V}  \in \mathbb{R}^{(d\times P) \times C_v}$, with $M =(d\times P)$ and $C=C_v$. For the text IAM, $\mathbf{F}\in \mathbb{R}^{M \times C}$ signifies $\mathbf{F_T} \in \mathbb{R}^{m \times C_t}$, where $M=m$ and $C=C_t$.

The base embedding $\mathbf{A}^{base}$ is designed to capture the essential semantic information of the training data and serves as an optimized initial value of the iterative approximation process during inference. Subsequently, inspired by the design of attention mechanism~\cite{attention}, which models the relationship of features via similarity, we apply the cross-attention module $\text{CA}_\text{I}$ to execute the approximation taking $\mathbf{A}^{base}$ as the Query and $\mathbf{F}$ as the Key and Value. To align the different dimensions necessary for executing the cross-attention between $\mathbf{F}$ and $\mathbf{A}^{base}$, and to ensure that the output maintains the same dimensionality as $\mathbf{A}^{base}$, we implement linear projection layers $\text{Q}_\text{I}$, $\text{K}_\text{I}$, and $\text{V}_\text{I}$, which project all feature dimensions to $D$. We also add a self-attention block $\text{SA}_\text{I}$ following the $\text{CA}_\text{I}$ to enhance the data approximation. Notably, longer videos and texts, which inherently contain more semantic information, pose greater challenges for effective approximation. Therefore, we repeat the approximation operations with $\text{CA}_\text{I}$ and $\text{SA}_\text{I}$, adjusting the iteration number $\#iter$ in accordance with the content length, with longer videos and texts assigned a greater $\#iter$. Through adaptive approximation iterated for $\#iter$ times, the resultant $\mathbf{A}^{\#iter}$ closely approximates $\mathbf{F}$, preserving its semantic information in a compact, low-dimensional space.



Meanwhile, inspired by~\cite{unroll1,unroll2,perceiver,unroll4,unroll5} that study the unrolling of the iterative algorithm and indicate that more unrolled iterations lead to higher model performance but lower computation efficiency, we unroll the iterative approximation process one time while assigning remaining iterations to the original iterative block denoted by $I$ to achieve good trade-offs between model performance and training efficiency. For clarity, we represent the cross-attention, self-attention, linear projection layers of the Query, Key, Value, and the output of the unrolled iteration as $\text{CA}$, $\text{SA}$, $\text{Q}$, $\text{K}$, $\text{V}$, and $\mathbf{A}^0$, respectively, as illustrated in Figure~\ref{fig2}, where the output $\mathbf{A}^0$ serves as the base embedding of the following iterative approximation.

IAM enables a single model to handle variable-length videos and texts by modulating $\#iter$, thereby maintaining a fixed low-dimensional output space for cross-modal alignment. Additionally, the output embedding $\mathbf{A}^{\#iter}\in \mathbb{R}^{N\times D}$ of the IAM is not only more compact than dense features in size but also retains a high fidelity approximation of the original dense feature $\mathbf{F}\in \mathbb{R}^{M\times C}$, leading to improved model performance on downstream tasks after alignment.

\noindent\textbf{Cross-modal Alignment:}
Given a batch (batch size $B$) of video and text pairs, IAMs produce paired embeddings $\mathbf{A}_\mathbf{V}^{\#iter} ang \mathbf{A}_\mathbf{T}^{\#iter} \in \mathbb{R}^{B \times N \times D}$ after the iterative approximation. Since $\mathbf{A}_\mathbf{V}^{\#iter}$ and $\mathbf{A}_\mathbf{T}^{\#iter}$ have the same size regardless of the input granularity, conventional contrastive learning~\cite{clip} can be directly applied to build alignments. We process paired embeddings with average pooling along dimension $N$, yielding vectorized features $\mathbf{\overline{\mathbf{A}}_\mathbf{V}}$ and $\mathbf{\overline{\mathbf{A}}_\mathbf{T}} \in \mathbb{R}^{B \times D}$ to minimize the Video-to-Text Contrastive (VTC) loss $\mathcal{L}_\text{VTC} = \frac{1}{2}(\mathcal{L}_\text{v2t}+\mathcal{L}_\text{t2v})$, which encompasses both video-to-text and text-to-video losses:

{\small
\vspace{-5mm}
\begin{align*}
    \mathcal{L}_\text{v2t} &= - \mathbb{E}_{\mathbf{v}} \Big[\log \frac{f(\mathbf{v}, \mathbf{t})}{\sum_{\mathbf{t}' \in \mathbf{\overline{\mathbf{A}}_\mathbf{T}}} f(\mathbf{v}, \mathbf{t'})} \Big], \\ \mathcal{L}_\text{t2v} &= - \mathbb{E}_{\mathbf{t}} \Big[\log \frac{f(\mathbf{v}, \ \mathbf{t})}{\sum_{\mathbf{v}' \in \mathbf{\overline{\mathbf{A}}_\mathbf{V}}} f(\mathbf{v}', \mathbf{t})} \Big],
\end{align*}
}\vspace{-3mm}

\noindent where $f(\mathbf{v},\mathbf{t}) = e^{s(\mathbf{v}, \mathbf{t})/\tau}$. $s$ is the cosine similarity and $\tau$ is a learnable scaling parameter. This loss function aims to maximize the cosine similarity between matching video-text pairs while minimizing it for non-matching pairs.

\section{Experiments}
\label{sec:experiment}
\subsection{Experimental Setups}
\noindent\textbf{Multi-grained Pretraining Data:}
We utilize the InternVid-10M-FLT~\cite{internvid} as the base pretraining dataset which consists of ten million short video clips collected from four million YouTube videos. Based on the statistics in \cite{internvid}, we treat video-text pairs in this dataset as single granularity, labeled short-video-short-text (SVST). In alignment with the data distribution of targeted downstream tasks, we choose to expand this base dataset with two additional granularities: long-video-long-text (LVLT) and long-video-short-text (LVST) pairs. More specifically, we perform Video and Text Integration ($\oplus_v$ and $\oplus_t$) through video concatenation and text concatenation on 4.5 million SVST pairs to build 1 million LVLT video-text pairs. These 4.5 million SVST pairs are selected by their source video IDs and timestamps, excluding those with fewer than four SVST pairs. The concatenation is performed only on videos sharing identical source IDs in time order, along with their corresponding texts. Subsequently, we apply the Text Compression ($\Theta_t$) through text summarization on the 1 million generated long texts using a self-hosted large language model (LLM) (Vicuna 13b-v1.5~\cite{vicuna,dhp}) to build 1 million LVST pairs. As demonstrated in Appendix B, the summarized short texts are shown to be semantically similar to the original long texts. Further details on the selection of the LLM and the potential use of Video Compression ($\Theta_v$) are provided in Appendix A and F.


\noindent\textbf{Model Pretraining:}
During the pretraining phase, we initialize our video and text feature encoder with CLIP weights~\cite{clip} (VIT-B/32 model). The pretraining is conducted on 32 NVIDIA Tesla-V100 GPUs and takes approximately 2.5 days, covering 5 epochs on our multi-grained dataset. We employ the AdamW optimizer~\cite{adamw}, using a batch size of 512. The learning rate for the video and text encoders is set to $1e^{-6}$, while for other components it is set to $1e^{-4}$, accompanied by a cosine learning rate schedule~\cite{cosine}. We process $d=16$ frames for short videos and $d=32$ frames for longer ones and empirically set $\#iter=1$ for short videos/texts and $\#iter=3$ for long videos/texts. The ablation on the $\#iter$ setups is in Section~\ref{sec:ablation}. $\tau$ is initially set to $0.07$~\cite{clip}.

\begin{table*}[t]
\footnotesize
\centering
\caption{Text-to-video (T2V) and Video-to-text (V2T) retrieval results using Recall at K (R@K) and Median Rank (MdR) metrics across different setups for ActivityNet Captions, MSR-VTT, and LSMDC datasets. Results for models that are not directly comparable are shown in gray. \#PT represents the size of the pretraining dataset.*Best among four setups. **No auxiliary captions.}
\vspace{-1mm}
\scalebox{0.72}{
\begin{tabular}{@{}llll|cccc|cccc|cccc@{}}
\hlineB{3}
\multirow{2}{*}{Text-to-video (T2V) Setup} & \multirow{2}{*}{Method} &  \multirow{2}{*}{\#PT} &  \multirow{2}{*}{Backbone}    & \multicolumn{4}{c|}{ActivityNet Captions T2V} & \multicolumn{4}{c|}{MSR-VTT T2V} &\multicolumn{4}{c}{LSMDC T2V} \\

                     &   &  &  & R@1   & R@5  & R@10   & MdR  & R@1     & R@5    & R@10 & MdR  & R@1    & R@5  & R@10   & MdR  \\ \hlineB{2}
 
\multirow{6}{*}{Zero-shot} & Frozen\cite{frozen}                &5.5M & TS4mer       &-      &-      &-      &-      & 24.7  & 46.9  & 57.2  &3.0    &13.6   & 27.9  & 33.5  & 32.0  \\
&VIOLET \cite{violet}   &185.8M & Swin$_{B}$     &-      &-      &-      &-      & 25.9  & 49.5  & 59.7  &-    &-   & -  & -  & - \\  
&CLIP4Clip \cite{clip4clip}             &136M & ViT$_{B32}$     &-      &-      &-      &-      & 32.0  & 57.0  & 66.9  &4.0    &15.1   & 28.5  & 36.4  & 28.0  \\  

&LocVTP \cite{locvtp}       & 136M   & ViT$_{B16}$  & -  & -  & -  & -   &32.7   &55.7   &64.9   & 4.0   &-   &-   &-   &-     \\

&ViCLIP\cite{internvid}             & 10M & ViT$_{L14}$  & 15.1  &-      &-      &-      & 24.0  &   -   & -     &-      &20.1   & -     & -     & -     \\ 
&Ours                          & 10M & ViT$_{B32}$  & 35.3  &64.3   &76.1   &3.0    & 37.3  & 63.8  & 72.5  &3.0    &12.9   & 28.3  & 35.1  & 35.5  \\

\rowcolor{gray!10}
&VideoPrism \cite{videoprism} & 618M & ViViT$_{B}$     &49.6      &76.7      &-      &-      & 51.4  & 74.4  & -  &-    &-   & -  & -  & - \\  \hlineB{2}
\multirow{3}{*}{Finetuned} & CLIPBert\cite{clipbert}            & 5.6M& Res$_{50}$   & 21.3  & 49.0  & 63.5  & 6.0   & -     & -     & -     &-      & -     & -     & -     & -     \\
&Perceiver-VL\cite{perceivervl}           & 5.5M& ViT$_{B32}$  & 33.9  & 62.1  & 76.4  & -     & 32.6  & 62.1  & 71.6  &-      &15.8   & 37.6  & 40.1  & -     \\

\multirow{3}{*}{Model with Similar FLOPs}&CLIP4Clip*~\cite{clip4clip}        &136M & ViT$_{B32}$  & 40.5  & 72.4  & -     & 2.0   & 44.5  & 71.5  & 81.6  &2.0    &\textbf{22.6}   & \textbf{41.8}  & 49.8  & 11.0  \\

&CLIP-ViP**~\cite{clipvip}       & 136M   & ViT$_{B32}$  & -  & -  & -  & -   &46.5   &\textbf{72.1}   &\textbf{82.5}   &-    &-   &-   &-   &-     \\

&Ours                      & 10M & ViT$_{B32}$  & \textbf{45.3} & \textbf{76.5} & \textbf{86.6} & \textbf{2.0} & \textbf{46.8} & 72.0 & 81.5  & \textbf{2.0} &21.2 &40.9 &\textbf{51.4} &\textbf{9.0} \\ \hlineB{2}


\rowcolor{gray!10}
\multirow{3}{*}{Finetuned} & Frozen\cite{frozen}                &5.5M & TS4mer       & 28.8  & 60.9  & -     & 3.0   &32.5   &61.5   &71.2   &3.0    &15.0   &30.8   &40.3   &20.0    \\

\rowcolor{gray!10}
Finetuned &X-CLIP\cite{xclip}                 & -   & ViT$_{B32}$  & 44.3  & 74.1  & -     & -     &46.1   &73.0   &83.1   &2.0    &23.3   &43.0   &-      &-      \\

\rowcolor{gray!10}
&PromptSwitch \cite{promptswitch}                 & -   & ViT$_{B32}$  & -  & -  & -     & -     &47.8   &73.9   &82.2   &-    &23.1   &41.7   &50.5      &-       \\

\rowcolor{gray!10}
&UCOFIA \cite{ucofia}                 & -   & ViT$_{B32}$  & 45.7  & 76.0  & -     & -     &49.4   &72.1   &-   &-    &-   &-   &-      &-       \\

\rowcolor{gray!10}
Retrieval-specific Model &X-CLIP \cite{xclip}                 & -   & ViT$_{B16}$  & 46.2  & 75.5  & -     & -     &49.3   &75.8   &84.8   &2.0    &26.1   &48.4   &-      &-       \\

\rowcolor{gray!10}
&CenterCLIP \cite{centerclip}          & -   & ViT$_{B16}$  & 46.2  & 77.0  & 87.6  & 2.0   &48.4   &73.8   &82.0   &2.0    &24.2   &46.2   &55.9   &8.0     \\ \hlineB{1}

\rowcolor{gray!10}
& TACo\cite{taco}                    & 136M& Res$_{152}$  & 30.4  & 61.2  &-      & 3.0   & 24.5  & 52.8  & 65.5  &5.0    & -     & -     & -     & -     \\
\rowcolor{gray!10}
& LF-VILA\cite{lfvila}               & 60M & Swin$_B$     & 35.3  & 65.4  & -     & 3.0   &-      &-      &-      &-      &-      &-      &-      &-       \\
\rowcolor{gray!10}
Finetuned & VIOLET \cite{violet}   &185.8M & Swin$_{B}$     &-      &-      &-      &-      & 34.5  & 63.0  & 73.4  &-    &16.1   & 36.6  & 41.2  & - \\  

\rowcolor{gray!10}
& LAVENDER \cite{lavender}   &30M & Swin$_{B}$     &-      &-      &-      &-      & 40.7  & 66.9  & 77.6  &-    &26.1  & 46.4  & 57.3  & - \\  

\rowcolor{gray!10}
Model with Higher FLOPs & HiTeA \cite{hitea}                  & 5M  & MViT$_B$    & 45.1   & 73.5  & 84.2  & -     &44.4   &69.3   &78.9   &-      &27.1   &46.2   &54.5   &-    \\

\rowcolor{gray!10}
& LocVTP \cite{locvtp}       & 136M   & ViT$_{B16}$  & -  & -  & -  & -   &46.3   &72.8   &82.0   &-    &-   &-   &-   &-     \\


\rowcolor{gray!10}
& ViCLIP \cite{internvid}            & 10M & ViT$_{L14}$ & 49.8   & -     & -     & -     &52.5   &-      &-      &-      &33.0   &-      &-      &-\\

\hline
\hline

\multirow{2}{*}{Video-to-text (V2T) Setup} & \multirow{2}{*}{Method} &  \multirow{2}{*}{\#PT} &  \multirow{2}{*}{Backbone}    & \multicolumn{4}{c|}{ActivityNet Captions V2T} & \multicolumn{4}{c|}{MSR-VTT V2T} &\multicolumn{4}{c}{LSMDC V2T} \\

                       & &  &  & R@1   & R@5  & R@10   & MdR  & R@1     & R@5    & R@10 & MdR  & R@1    & R@5  & R@10   & MdR  \\ \hlineB{2}
 
\multirow{2}{*}{Zero-shot} & ViCLIP\cite{internvid}             & 10M & ViT$_{L14}$  & 24.0  &-      &-      &-      & 41.3  &   -   & -     &-      &16.9   & -     & -     & -     \\ 

&Ours                       & 10M & ViT$_{B32}$  & 35.0  &64.9   &77.6   &3.0    & 39.5  & 64.6  & 73.0  &2.0    &13.0   & 26.8  & 33.9  & 43.0  \\

\rowcolor{gray!10}
&VideoPrism~\cite{videoprism}                 & 618M & ViViT$_B$  & 47.9  &75.0  &-   &-    & 50.2  & 73.2  & -  &-    &-   & -  & -  & -  \\ \hlineB{2}
Finetuned & CLIP4Clip*~\cite{clip4clip}       &136M & ViT$_{B32}$  & 42.5  & 74.1  & 85.8  & 2.0   & 43.1  & 70.5  & 81.2  &2.0    &20.9   & \textbf{40.7}  & 49.1  & 11.0  \\

Model with Similar FLOPs & Ours                             & 10M & ViT$_{B32}$  & \textbf{45.0} & \textbf{76.4} & \textbf{87.3} & \textbf{2.0} & \textbf{47.4} & \textbf{74.0} & \textbf{81.8}  & \textbf{2.0} &\textbf{21.9} &40.5 &\textbf{50.2} &\textbf{10.0} \\ \hlineB{2}


\rowcolor{gray!10}
&X-CLIP\cite{xclip}                 & -   & ViT$_{B32}$  & 43.9  & 73.9  & -     & -     &46.8   &73.3   &84.0   &2.0    &22.5   &42.2   &-      &-      \\

\rowcolor{gray!10}
Finetuned &PromptSwitch \cite{promptswitch}                 & -   & ViT$_{B32}$  & -  & -  & -     & -     &46.0   &74.3   &84.8   &-    &22.0   &40.8   &50.3      &-       \\

\rowcolor{gray!10}
&UCOFIA\cite{ucofia}                 & -   & ViT$_{B32}$  & 46.3  & 76.5  & -     & -     &47.1   &74.3   &-   &-    &-   &-   &-      &-      \\

\rowcolor{gray!10}
Retrieval-specific Model & X-CLIP\cite{xclip}                 & -   & ViT$_{B16}$  & 46.4  & 75.9  & -     & -     &48.9   &76.8   &84.5   &2.0    &26.9   &46.2   &-      &-       \\

\rowcolor{gray!10}
&CenterCLIP \cite{centerclip}       & -   & ViT$_{B16}$  & 46.7  & 77.1  & 88.0  & 2.0   &47.7   &75.0   &83.3   &2.0    &24.5   &46.4   &55.8   &7.0     \\ \hlineB{1}


\rowcolor{gray!10}
Finetuned, Model with Higher FLOPs& ViCLIP \cite{internvid}            & 10M & ViT$_{L14}$  & 48.1  & -     & -     & -     &51.8   &-      &-      &-      &32.5   &-      &-      &-\\ \hlineB{3}
\end{tabular}
}

\label{t2v}
\end{table*}

\subsection{Cross-modal Retrieval}
\noindent\textbf{Setting:} 
We finetune the pretrained model on ActivityNet Captions (paragraph-to-video)~\cite{krishna2017dense}, MSR-VTT (9k)~\cite{msrvtt, msrvtt9k}, and LSMDC~\cite{lsmdc} datasets for text-to-video and video-to-text retrieval tasks. Following previous work~\cite{clip4clip, internvid}, we report both zero-shot and finetuning retrieval performance with recall@K (R@K) and median rank (MdR). We finetune our pretrained model for 50 epochs on 8 NVIDIA Tesla-V100 GPUs. We adopt an AdamW~\cite{adamw} optimizer~\cite{adamw} and set the learning rate of the video and text encoders to $1e^{-7}$, and the other part $1e^{-5}$ with a cosine annealing scheduler~\cite{cosine}. Based on the average video durations and text length of the datasets, we empirically determine MSR-VTT and LSMDC as SVST,  ActivityNet Captions as LVLT and set the $\#iter$ accordingly (see statistics and the $\#iter$ setups in Appendix E). 

\noindent\textbf{Retrieval Results:}
Due to the complexity of training strategies and model backbones design, we present our comparisons from multiple perspectives, as outlined in Table~\ref{t2v}. 

Compared with models of \textbf{using the same ViT-B/32 backbone or having similar FLOPs}, our approach outperforms other baselines~\cite{perceivervl,clip4clip,clipvip} on the long video ActivityNet Captions dataset. PerceiverVL~\cite{perceivervl} uses a similar iterative module, but our approach achieves T2V R@1 gains of +11.4\% on ActivityNet, +14.2\% on MSR-VTT, and +5.4\% on LSMDC. Notably, in zero-shot settings, our approach surpasses the finetuned PerceiverVL on ActivityNet (R@1 +1.4\%) and MSR-VTT (R@1 +4.7\%), highlighting the superior multi-grained data modeling capability of our IAM module. Compared to CLIP-based models, our model significantly outperforms CLIP4Clip~\cite{clip4clip} on ActivityNet (T2V/V2T R@1 +4.8\%/+2.5\%). Although pre-trained on a dataset (10M) which is much smaller than CLIP4Clip~\cite{clip4clip} and CLIP-ViP~\cite{clipvip} (136M), our model performs comparably on the short video MSR-VTT and LSMDC, demonstrating the effectiveness of our approach in expanding data granularity and aligning multi-grained features.

Among \textbf{retrieval-specific models}, which incorporate specialized retrieval structures or strategies, our GEXIA as a general pretraining model, also outperforms X-CLIP$_{B32}$~\cite{xclip} on two datasets (T2V/V2T R@1 +1.0\%/+1.1\% on ActivityNet and +0.7\%/+0.6\% on MSR-VTT). Furthermore, it achieves comparable results with PromptSwitch~\cite{promptswitch} and UCOFIA~\cite{ucofia}, emphasizing the strengths of our GEX data preprocessing method and IAM module design. Additionally, our method demonstrates low retrieval complexity and high inference efficiency, as detailed in Appendix C.

Although models with \textbf{higher computational demands (higher FLOPs)} tend to exhibit better performance, our approach often achieves comparable, and in some cases superior results. For instance, it outperforms previous works such as VIOLET~\cite{violet} (T2V R@1 +12.3\% on MSR-VTT and +5.1\% on LSMDC) and LAVENDER~\cite{lavender} (T2V R@1 +6.1\% on MSR-VTT), both of which use additional masked language modeling pretraining (MLM). Our model also surpasses LF-VILA~\cite{lfvila} (T2V R@1 +10.0\% on ActivityNet) and HiTeA~\cite{hitea} (T2V R@1 +0.2\% on ActivityNet and +2.4\% on MSR-VTT), which are specifically designed for long-form and hierarchical video-text alignment. These achievements highlight the critical role of the IAM in effectively aligning multi-grained video-text pairs.


In the \textbf{zero-shot setting}, our method significantly outperforms ViCLIP~\cite{internvid} on the long-video ActivityNet Captions. ViCLIP employs a ViT$_{L14}$ backbone, which demands 17$\times$ more computational resources (see Appendix D). Despite this disparity, our model achieves remarkable gains of +20.2\%/+11.0\% in T2V/V2T R@1, underscoring the effectiveness of the GEX method in enhancing the model during pretraining, especially for long-video datasets. Furthermore, our approach surpasses other methods in T2V retrieval on MSR-VTT, such as CLIP4Clip~\cite{clip4clip} and LocVTP~\cite{locvtp}, both of which were pretrained on the large-scale HowTo100M~\cite{howto100m} datasets (136M video-text pairs). While CLIP4Clip outperforms our model in LSMDC zero-shot retrieval due to the closer domain similarity between LSMDC (average clip length: 4.7 seconds) and HowTo100M (average clip length: 4 seconds, compared to our dataset’s 10 seconds per clip), our model still achieves competitive performance after fine-tuning. We also include VideoPrism~\cite{videoprism} for reference but do not make direct comparisons, as it leverages a massive (618M) non-public dataset and uses a ViViT backbone, which is 30$\times$ more computationally expensive than our approach (see Appendix D).



\begin{table}[t]
\centering
\footnotesize

\caption{Classification results on the LVU, COIN, Charades-Ego, and How2QA datasets. Results from models using larger backbones on the LVU and COIN datasets are shown in gray for reference. \#PT denotes the size of the pretraining dataset, ZS indicates zero-shot results, and FT refers to finetuned results.}
\vspace{-2mm}

 \subfloat[Accuracy Results on LVU dataset.]
    {
    \scalebox{0.59}{
        \begin{tabular}{lccc}
        \hlineB{3}
        Method             & Relation & Speaking & Scene \\ \cline{1-4}
        VideoBERT \cite{videobert}          & 52.4         & 37.9            & 54.9  \\
        Obj.T4mer \cite{lvu}                & 53.1         & 39.4            & 56.9  \\
        LST\cite{vis4mer}                   & 52.4         & 37.3            & 62.8  \\
        Orthoformer \cite{orthoformer}      & 50.0         & 38.3            & 66.3  \\
        ViS4mer\cite{vis4mer}               & 57.1         & 40.8            & 67.4  \\

         Ours            & \textbf{61.9}         & \textbf{42.7}            & \textbf{70.9}  \\ \hline

        \rowcolor{gray!10}
        LF-VILA\cite{lfvila}            & 61.5         & 41.3            & 68.0  \\
        
         \rowcolor{gray!10}S5$_\text{60frame}$\cite{wang2023selective}        & 61.9        & 41.8           & 69.9 \\
        
        \rowcolor{gray!10}
        S5$_\text{100frame}$\cite{wang2023selective}        & 66.7        & 41.8           & 73.3 \\ 
        
        \rowcolor{gray!10}
        MA-LMM$_\text{100frame}$\cite{mallm}        & 58.2        & 44.8           & 80.3 \\ \hlineB{3}
        
        \end{tabular}
        }
        \label{lvu_table}
    }\hfill
 \subfloat[Results on COIN dataset.]
    {
    \scalebox{0.59}{
        \begin{tabular}{lcc}
        \hlineB{3}
        Method    & \#PT   & Accuracy \\ \hline
        CLIPBert \cite{clipbert} &  5.6M  & 65.4     \\
        MIL-NCE \cite{milnce}    & 136M  & 70.2      \\
        SlowFast \cite{slowfast} &  136M & 71.6     \\
        VideoCLIP \cite{videoclip} &  136M    & 72.5 \\
        Ours        &  10M   & \textbf{84.8}     \\ \hline
        \rowcolor{gray!10} TS4mer \cite{timesformer} & 0.37M & 83.5  \\
        \rowcolor{gray!10} TS4mer \cite{timesformer} & 136M & 85.3  \\
        \rowcolor{gray!10} LF-VILA \cite{lfvila}    &  60M & 85.7     \\ 
        \rowcolor{gray!10} S5$_\text{100frame}$ \cite{s5}    &  60M & 90.4     \\
        \rowcolor{gray!10} MA-LMM$_\text{100frame}$ \cite{mallm}    &  - & 93.2     \\
        \hlineB{3}
        \end{tabular}
        }
        \label{coin_table}
    }\hfill
 \subfloat[Results on Charades-Ego dataset.]
    {
    \scalebox{0.66}{
        \begin{tabular}{lccc}
        \hlineB{3}
        Method  & ZS mAP & FT mAP & $\Delta_{\text{FT-ZS}}$ \\ \hline
        Actor \cite{actor}             & -     & 20.0 &-\\
        SSDA \cite{ssda}               & -     & 23.1 &-\\
        I3D \cite{ssda}                & -     & 25.8 &-\\
        Ego-Exo \cite{egoexo}                       & -     & 30.1 &-\\
        EgoVLP \cite{egovlp}           & 25.0  & 32.1 &7.1\\
        HierVL-Avg \cite{hiervl}       & 25.2  & 32.6 &7.4 \\ 
        HierVL-SA \cite{hiervl}        & \textbf{26.0}  & \textbf{33.8} &7.8 \\ 
        Ours                          & 23.4  & \textbf{33.8} &\textbf{10.4} \\ \hlineB{3}
        \end{tabular}
        }
        \label{charades_ego_table}
    }\hfill
\subfloat[Results on How2QA dataset.]
    {
    \scalebox{0.66}{
        \begin{tabular}{lcc}
        \hlineB{3}
        Method    & \#PT   & Accuracy \\ \hline
        ATP \cite{atp} &  400M  & 65.4     \\
        Hero \cite{hero} &  136M & 71.6     \\
        LF-VILA \cite{lfvila}    & 60M  & 76.1     \\
        FrozenBiLM \cite{FrozenBiLMzero} & 10M & 81.5 \\
        
        SeViLA \cite{sevila} &  -    & 83.7 \\
        SiaSamRea \cite{siasamrea} & 5.6M & 84.1 \\
        JustAsk \cite{justask} &  69M & 84.4\\
        Ours        &  10M   & \textbf{84.8}     \\ \hlineB{3}
        \end{tabular}
        }
        \label{how2qa_table}
    }\hfill
\vspace{-5mm}
\label{video_class}
\end{table}

\subsection{Classification}
\noindent\textbf{Setting:}
We evaluate our GEXIA method on video-related classification tasks using the LVU~\cite{lvu}, COIN~\cite{coin}, Charades-Ego~\cite{charades}, and How2QA~\cite{hero} datasets. LVU and COIN are designed for long-form video understanding. Charades-Ego, an action-classification-targeted dataset with egocentric videos, is used to evaluate the domain transfer performance of our approach.  How2QA is leveraged to evaluate our approach on a cross-modal classification task.

Following the methodology of LF-VILA \cite{lfvila}, we exclusively employ the video branch and add a fully-connected layer for classification on the LVU and COIN datasets reporting the results in terms of accuracy. The $\#iter$ is set to 3. We select the three content understanding tasks from the LVU dataset (Relationship, Way of Speaking, and Scene) as outlined in LF-VILA. For the Charades-Ego dataset, we replicate the process used by HierVL~\cite{hiervl}, treating it as a retrieval task across 157 distinct short action captions, and evaluate using the mean average precision (mAP) metric, with $\#iter$ configured per LVST settings. How2QA is a multi-choice video question answering (VQA) dataset. It consists of 44K short QA pairs and each question has one correct answer and 3 wrong answers. We concatenate questions and answers as text data. For each downstream task, our pretrained model was finetuned for 100 epochs on 8 NVIDIA Tesla-V100 GPUs. We utilized an AdamW optimizer~\cite{adamw}, setting the learning rate for the video and text encoders at $1e^{-7}$, and for other components at $1e^{-6}$ managed by a cosine annealing schedule~\cite{cosine}. Detailed $\#iter$ settings are provided in Appendix E.

\noindent\textbf{Long-form Video Understanding:} Our GEXIA method achieves state-of-the-art performance in video classification tasks \textbf{on the LVU dataset}, as shown in Table~\ref{lvu_table}, even outperforming some models that may have been pretrained on larger datasets or utilize backbones with higher FLOPs. Notably, while the recent work S5$_\text{60frame}$~\cite{s5} is specifically designed for long-form videos with state compression, our model still demonstrates superior performance (Accuracy +0.9\% for Speaking, +1.0\% for Scene), validating the success of our proposed IAM module. Furthermore, our model surpasses LF-VILA~\cite{lfvila}, which is also designed for long-form video pretraining, by a considerable margin (Accuracy +0.4\% for Relationship, +1.4\% for Speaking, +1.9\% for Scene). Although the MA-LMM method~\cite{mallm} uses a pretrained large multimodal model and achieves higher results on the Speaking and Scene labels, our model still outperforms it on the Relationship label. Our model also attains the highest score \textbf{on the COIN dataset} among comparable methods~\cite{milnce, videobert, slowfast, clipbert}. The results, shown in Table~\ref{coin_table}, are competitive with LF-VILA~\cite{lfvila} and TimeSformer~\cite{timesformer}, even though those models are pretrained on more data or utilize backbones with higher FLOPs.

These findings highlight the effectiveness of our video branch model in capturing long-form temporal information. Our model's significant advantage stems from its ability to handle the varying information density across the temporal dimension in long-form videos. This adaptability to different granularities is a key factor in our model's success.


\noindent\textbf{Domain Transfer to Egocentric Video:}
We present the results from the Charades-Ego dataset in Table~\ref{charades_ego_table} to demonstrate the robustness of our model. 
While our model shows relatively lower zero-shot results compared to baseline methods due to the domain gap between our pretrained data and the egocentric evaluation data, it significantly outperforms other baselines after finetuning. Notably, it achieves performance comparable to the HierVL~\cite{hiervl} method, which is pretrained on the in-domain Ego4D dataset~\cite{ego4d}. The key observation is the substantial disparity between our model's zero-shot and finetuned mAP results, highlighting its ability to adapt effectively to different video content after finetuning, even when pretrained on an out-of-domain dataset.

\noindent\textbf{Cross-modal Classification:}
We finally show results on the  How2QA dataset in Table~\ref{how2qa_table}. 
Our model outperforms the baseline methods, achieving the highest accuracy. Our result significantly surpasses the video-text pretraining models~\cite{hero, lfvila}, and also outperforms the method leveraging LLMs~\cite{sevila}, as well as those specifically designed for VQA tasks~\cite{siasamrea, justask}. The superior performance demonstrates that the low-dimensional embeddings of multi-grained data generated by IAM approximate dense features well without losing critical details in videos and texts.

\subsection{Ablation Studies}
\label{sec:ablation}
We conduct ablation studies to investigate different settings of GEX and IAM, especially pretraining data compositions and iteration numbers ($\#iter$). We also study the setups of the encoders, the random Integration in GEX, and the selections of LLM in GEX detailed in Appendix A.

\begin{table}[!t]
\centering
\footnotesize
\caption{Retrieval results of finetuned model on ActivityNet Captions dataset across different pretraining data setups.}
\vspace{-3mm}
\scalebox{0.8}{
        \begin{tabular}{l|ccc|ccc}
        \hlineB{3}
        \multirow{2}{*}{Data Setup} & \multicolumn{3}{c|}{Text-to-video} & \multicolumn{3}{c}{Video-to-text}\\
                                 & R@1 & R@5 & R@10 & R@1 & R@5 & R@10  \\ \hline
        short-video-short-text (SVST)                                                                          &41.6     & 72.8     & 83.7      & 41.2     & 72.6     & 84.0       \\ 
        +long-video-long-text (LVLT)                                                                   & 42.8     & 72.8     & 84.0      & 41.8     & 73.0     & 84.3         \\
        +long-video-short-text (LVST)                                                             & \textbf{43.1}     & \textbf{73.2}     & \textbf{84.5}     & \textbf{42.6}     & \textbf{73.4}     & \textbf{84.8}        \\ \hlineB{3}
        \end{tabular}
        }
\vspace{-5mm}
\label{ablation_data}
\end{table}

\noindent\textbf{Pretraining Data Compositions:}
To validate the performance improvement from introducing multi-grained data with GEX, we pretrain our model with different compositions of pretraining data and finetune the model on ActivityNet for retrieval. We use $\#iter$=1 to establish a fair comparison.  As shown in Table~\ref{ablation_data}, the data composition with three different granularities works best while the training without multi-grained data is the worst. This result demonstrates the effectiveness of the proposed GEX pipeline.

\begin{figure}[t]
    \centering
    \includegraphics[width=0.94\columnwidth]{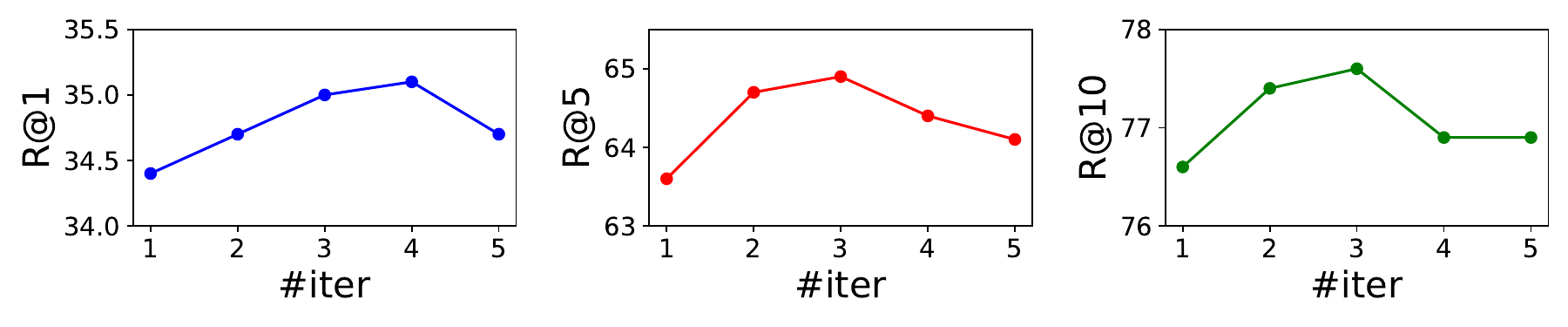}
    \vspace{-4mm}
    \caption{Zero-shot T2V retrieval results with different $\#iter$ for long video/text data on the ActivityNet Captions dataset.}
    \label{fig3}
    \vspace{-5mm}
\end{figure}

\noindent\textbf{Iteration Number Selection for Long Videos and Texts:} We empirically study the best option of the $\#iter$ for long videos and texts using ActivityNet Captions T2V retrieval task as a benchmark. As shown in Figure~\ref{fig3}, it is clear that, ranging from 1 to 5, $\#iter=3$ strikes the best balance between effectiveness and efficiency for long videos and texts. Therefore, we set $\#iter=3$ for long video/text data in the training and inference stage.

\noindent\textbf{Iteration Number in the Pretraining Stage:}
To fully understand the impact of iteration numbers in IAMs on multi-grained information extraction and alignments, we test different setups of video-text $\#iter$ in the pretraining stage. Given the optimal setup of $\#iter$ during pretraining, i.e., SVST: 1-1; LVLT: 3-3; LVST: 3-1, we introduce two suboptimal setups: fixed $\#iter$ to 1-1 or 3-3 across SVST, LVLT, and LVST regardless of their actual granularity. Subsequently, we conduct zero-shot retrieval experiments on ActivityNet Captions \cite{activitynet} and MSR-VTT\cite{msrvtt} datasets, aiming to evaluate the performance of the pre-trained models while isolating the results from any influences of finetuning. 

As shown in Table~\ref{ablation_iter_pretrain}, using a 3-3 only setup yields results that are comparable to the optimal setup on the short-video-short-text (SVST) MSR-VTT dataset while using a 1-1 only setup results in a large performance drop. This reveals the fact that assigning smaller $\#iter$ to short inputs is sufficient while assigning larger $\#iter$ to long inputs is necessary during pretraining to achieve better performance. For the long-video-long-text (LVLT) ActivityNet Captions dataset, the optimal setup outperforms both the 1-1 only and 3-3 only setups. This underscores the importance of proper $\#iter$ counts to the length of the input during pretraining, as this strategy is crucial for achieving the best results in information extraction and alignment with multi-grained data. Overall, we set the $\#iter=1$ for short videos/texts and $\#iter=3$ for long videos/texts.


\begin{table}[!t]
\centering
\caption{Ablation study results on different setups of $\#iter$.}
\vspace{-.3cm}
\footnotesize
    \subfloat[Zero-shot retrieval results across different \textbf{pretraining} $\#iter$ setups.]
    {
    \scalebox{0.69}{
        \begin{tabular}{l|ccc|ccc}
\hlineB{3}
\multirow{2}{*}{\begin{tabular}[c]{@{}l@{}}Pretraining Video-Text\\ $\#iter$ setup\end{tabular}} & \multicolumn{3}{c|}{ActivityNet Captions T2V} & \multicolumn{3}{c}{ActivityNet Captions V2T}\\
& R@1      & R@5      & R@10     & R@1      & R@5      & R@10  \\ \hlineB{1}
Only 1-1 & 35.2     & 63.9     & 75.6     & 34.4     & 63.6     & 76.6  \\ 
Only 3-3 & 35.0     & 64.0     & 76.2     & 34.0     & 64.1     & 76.4  \\
1-1\&3-1\&3-3  & \textbf{35.3}     & \textbf{64.3}     & \textbf{76.3}     & \textbf{35.0}     & \textbf{64.9}     & \textbf{77.6} \\ \hlineB{3}

\multirow{2}{*}{\begin{tabular}[c]{@{}l@{}}Pretraining Video-Text\\ $\#iter$ setup\end{tabular}} & \multicolumn{3}{c|}{MSRVTT T2V} & \multicolumn{3}{c}{MSRVTT V2T} \\
& R@1      & R@5      & R@10     & R@1      & R@5      & R@10   \\ \hlineB{1}
Only 1-1   &  36.6      & 63.5     & 72.3     & 39.8     & 63.5     & 72.3    \\ 
Only 3-3  & 36.7      & \textbf{64.8}     & 72.3     & \textbf{40.0}     & 64.3     & \textbf{73.4}  \\
1-1\&3-1\&3-3 & \textbf{37.3}      & 63.8     & \textbf{72.5}     & 39.5     & \textbf{64.6}     & 73.0  \\ \hlineB{3}

\end{tabular}
    } \label{ablation_iter_pretrain}
    }\hfill

\subfloat[Experiment results across different \textbf{inference} $\#iter$ setups. FT: Finetuned results; ZS: Zero-shot results.]
    {
    \scalebox{0.69}{
    \begin{tabular}{c|ccc|ccc}
\hlineB{3}
\multirow{2}{*}{\begin{tabular}[c]{@{}l@{}}Inference Video-Text\\ $\#iter$ setup\end{tabular}} & \multicolumn{3}{c|}{FT ActivityNet Captions T2V} &\multicolumn{3}{c}{ZS MSRVTT T2V} \\
 & R@1 & R@5 & R@10 & R@1 & R@5 & R@10  \\ \hline
1-1                   & 43.1     & 73.2     & 84.5     & \textbf{37.3} &\textbf{63.8} & \textbf{72.5} \\ 
3-1                     & 44.8     & 75.5     & 86.0   & 36.3 &62.5 & 72.1 \\       
3-3                     & \textbf{45.3}& \textbf{76.5}  & \textbf{86.6} & 36.3 &62.8 & 71.7 \\ 
\hlineB{3}
\multirow{2}{*}{\begin{tabular}[c]{@{}l@{}}Inference Video-Text\\ $\#iter$ setup\end{tabular}} & \multicolumn{3}{c|}{ZS LSMDC T2V} & \multicolumn{3}{c}{ZS Charades-Ego}\\
 & R@1 & R@5 & R@10 & &mAP  & \\ \hline
1-1                    &12.9 & \textbf{28.3} & \textbf{35.5} & &22.9 & \\ 
3-1                     &13.8 & 27.7 & 35.3 &   & \textbf{23.4}  &\\       
3-3                     &\textbf{14.1} & 26.6 & 33.5 & & 23.3 & \\ \hlineB{3}
\end{tabular}
} \label{ablation_iter_infer}
} 
    

\vspace{-3mm}
\end{table}
\noindent\textbf{Iteration Number in the Inference Stage:}
We explore the different iteration numbers and their impact in the inference stage after pretraining the model with multi-grained data pairs. As reported in Table~\ref{ablation_iter_infer}, it becomes evident that the number of iterations should be aligned with the form of the video and text content. Benchmarked against InternVid~\cite{internvid}, which comprises short videos (average 10 sec) with short texts (9.4 words), the ActivityNet Captions dataset — characterized by an average video length of 180 sec and an average text length of 49.2 words (see Appendix E) — falls into the long-video-long-text (LVLT) category. For the ActivityNet, the optimal iteration configuration is found to be video-text $\#iter$=3-3. In contrast, the MSR-VTT and LSMDC, which have average video lengths of 14.8 sec and 4.7 sec, and average text lengths of 9.3 and 9.7 words, respectively, fit into the short-video-short-text (SVST) category. For these datasets, the video-text $\#iter$=1-1 configuration yields the best results. The Charades-Ego, with an average video length of 31.5 sec and an average text length of 3.9 words, is categorized as a long-video-short-text (LVST) dataset. Here, the video-text $\#iter$=3-1 configuration is most effective.

We also visualize the video-text alignment with different setups in Appendix G. These findings not only affirm the versatility of our model in processing multi-grained data but also highlight its ability to adapt during the inference stage with just the modification of the $\#iter$ parameter. This adaptability is crucial for handling various types of video-text pairs, ensuring optimal performance across diverse downstream datasets. 







\section{Conclusion and Future Work}
\label{sec:conclusion}
In this paper, we propose GEXIA for scalable multi-grained video-text learning. We first introduce GEX, a video-text data granularity expansion method that transforms single-grained video-text datasets into a multi-grained format. We also present a multi-grained video-text alignment framework, which features the IAM specifically designed to embed and align multi-grained information effectively. Remarkably, our model achieves state-of-the-art or comparable results across seven downstream datasets, underscoring the effectiveness of both the GEX and IAM. In the future, we aim to develop a multi-grained video-text benchmark to evaluate our model and other baseline approaches comprehensively. We hope that the GEXIA framework will prove valuable to the broader research community.

\appendix
\setcounter{page}{1}

\section{Ablation Studies}

\noindent\textbf{Encoder Weight Setup:}
We perform an ablation study to explore different weight setups for the dense feature encoders prior to pretraining and present the finetuning retrieval results on the ActivityNet Captions dataset in Table~\ref{supple_encoder_weight}. These results demonstrate a significant improvement when initializing the feature backbone with CLIP pretraining weights as opposed to training from scratch. It's evident that the InternVid-10M-FLT dataset isn't large enough to independently train our model from scratch. The common practice of initializing the video feature backbone with pretrained weights from an image model~\cite{clip} works well for large-scale video-text pretraining. Furthermore, our observations indicate that freezing the encoders leads to inferior outcomes. The best way is to unfreeze the backbone with a smaller learning rate. This strategy suggests that the image representations learned from CLIP do not seamlessly transfer to video representations.

\noindent\textbf{Random Integration in GEX:}
We conduct an ablation experiment to further demonstrate that, even for single-grained pretraining datasets without prior information, using random GEX concatenation operations ($\oplus_v$ and $\oplus_t$) can still improve the model's performance. We replace Long-Video-Long-Text (LVLT) and Long-Video-Short-Text (LVST) pairs with the same number of randomly concatenated short video-text pairs. The rest of the setups remain consistent. As the results of T2V and V2T retrieval on ActivityNet Captions shown in Table~\ref{data_random}, random concatenation is suboptimal to the current data pipeline, leading to 0.5\%/1.1\% (T2V/V2T R@1) performance drop. Compared to the approach with SVST only, random concatenation still provides 3.2\%/2.7\% performance improvements, which can be viewed as an alternative way when the prior knowledge of videos is not available.

\begin{table}[!t]
\centering
\footnotesize
\caption{Retrieval results of finetuned model on ActivityNet Captions dataset across different encoder weight setups.}
 \scalebox{0.88}{
    \begin{tabular}{l|ccc|ccc}
    \hlineB{3}
    \multirow{2}{*}{Encoder Weight Setup} & \multicolumn{3}{c|}{Text-to-video} & \multicolumn{3}{c}{Video-to-text}\\
                         & R@1 & R@5 & R@10 & R@1 & R@5 & R@10  \\ \hline
Scratch                  & 24.7     & 51.1     & 64.1      & 24.7     & 52.6     & 65.9       \\ 
Freeze CLIP       & 37.5     & 67.4     & 80.6      & 36.9     & 68.0     & 80.9         \\                 
Unfreeze CLIP & \textbf{45.3}     & \textbf{76.5}     & \textbf{86.6}     & \textbf{45.0}     &      \textbf{76.4}     & \textbf{87.3}        \\
    \hlineB{3}
    \end{tabular}}
    \label{supple_encoder_weight}

\end{table}
\begin{table}[t]
    \centering
    \footnotesize
    \caption{Retrieval results of finetuned model on ActivityNet Captions dataset across different concatenation operations in GEX.}
    \scalebox{0.88}{
    \begin{tabular}{l|ccc|ccc}
        \hlineB{3}
        \multirow{2}{*}{Pretraining Data Pairs} & \multicolumn{3}{c|}{Text-to-video} & \multicolumn{3}{c}{Video-to-text}\\
          & R@1 &  R@5 &  R@10 & R@1 &  R@5 &  R@10 \\\hline
        SVST only & 41.6  & 72.8  & 83.7 & 41.2 & 72.6 & 84.0 \\
        Random Concat. & 44.8 & 75.1 & 86.1 & 43.9 & 74.7 & 85.5 \\
        Concat. w/ source IDs & \textbf{45.3}& \textbf{76.5} & \textbf{86.6} & \textbf{45.0} & \textbf{76.4} & \textbf{87.3} \\\hlineB{3}
    \end{tabular}}
    \label{data_random}
\end{table}
\begin{table*}[!h]
\centering
\footnotesize
\caption{LLM assessment results. RG: ROUGE score \cite{rouge}; BERT: BERTScore-F1 \cite{bertscore}; Time: Average running time on one summarization task. The model underlined is the final selected LLM.}
{
\scalebox{1}{
\begin{tabular}{l|cccc|c|l|cccc|c}
\hlineB{3}
LLM            & RG-1 & RG-2 & RG-L & \multicolumn{1}{l|}{BERT} & \multicolumn{1}{l|}{Time (s)} & LLM            & RG-1 & RG-2 & RG-L & \multicolumn{1}{l|}{BERT} & \multicolumn{1}{l}{Time (s)}  \\ \hline
Longchat 7b\cite{longchat}  & 0.43  & 0.20  & 0.34  & 0.90                            & 2.45    & Internlm 7b \cite{internlm}     & 0.32  & 0.13  & 0.26  & 0.87                            & 3.36      \\
OpenLlama 7b \cite{openllama}    & 0.48  & 0.24  & 0.39  & 0.89                            & 1.03   &  RWKV-4 7b \cite{rwkv}     & 0.28  & 0.11  & 0.23  & 0.88                            & 1.04      \\
OpenLlama 13b \cite{openllama}    & 0.38  & 0.18  & 0.31  & 0.90                            & 2.19  & Vicuna 7b \cite{vicuna}       & 0.46  & 0.23  & 0.37  & 0.91                            & 1.91                                                         \\
Fastchat t5 3b\cite{fastchatt5}  & 0.48  & 0.25  & 0.40  & 0.91                            & 2.97      &  \underline{Vicuna 13b} \cite{vicuna}      & 0.48  & 0.24  & 0.39  & 0.92                            & 2.19                 \\
Dolly-v2-7b \cite{dolly}         & 0.36  & 0.15  & 0.28  & 0.89                            & 4.93  & Vicuna 33b \cite{vicuna}     & 0.50  & 0.27  & 0.42  & 0.92                            & 7.38  \\ \hlineB{3}

\end{tabular}
}\label{supple_LLM}
}

\end{table*}

\noindent\textbf{LLM for Text Compression (Summarization) in GEX:}
We test the effectiveness of various open-source Large Language Models (LLMs) serving as the Text Compression ($\Theta_t$) operation to create summaries. This assessment is carried out through 100 summarization tasks, randomly selected from the LSMDC validation set, which comprises 362 short video clips. Using GPT4's summaries as references, we evaluate different models' performance based on the relevance score of their generated summaries, involving ROUGE score~\cite{rouge} and BERTScore~\cite{bertscore}, along with the average running time on one summarization instance as presented in Table~\ref{supple_LLM}. Vicuna 13b-v1.5~\cite{vicuna} emerges as the top performer, particularly in terms of the highest BERTScore, striking the best balance between performance and runtime. Note that we don't use GPT4 directly as the Text Compression operator due to its high cost and limitations in parallel processing capabilities.

\section{Semantic Alignment between Long and Summarized Texts}
To verify that the summarized short texts retain semantic similarity with the original concatenated long texts, ensuring consistency between them, we analyze the cosine similarity between the long and summarized short text features.

\begin{figure}[t]
    \centering
    \includegraphics[width=\columnwidth]{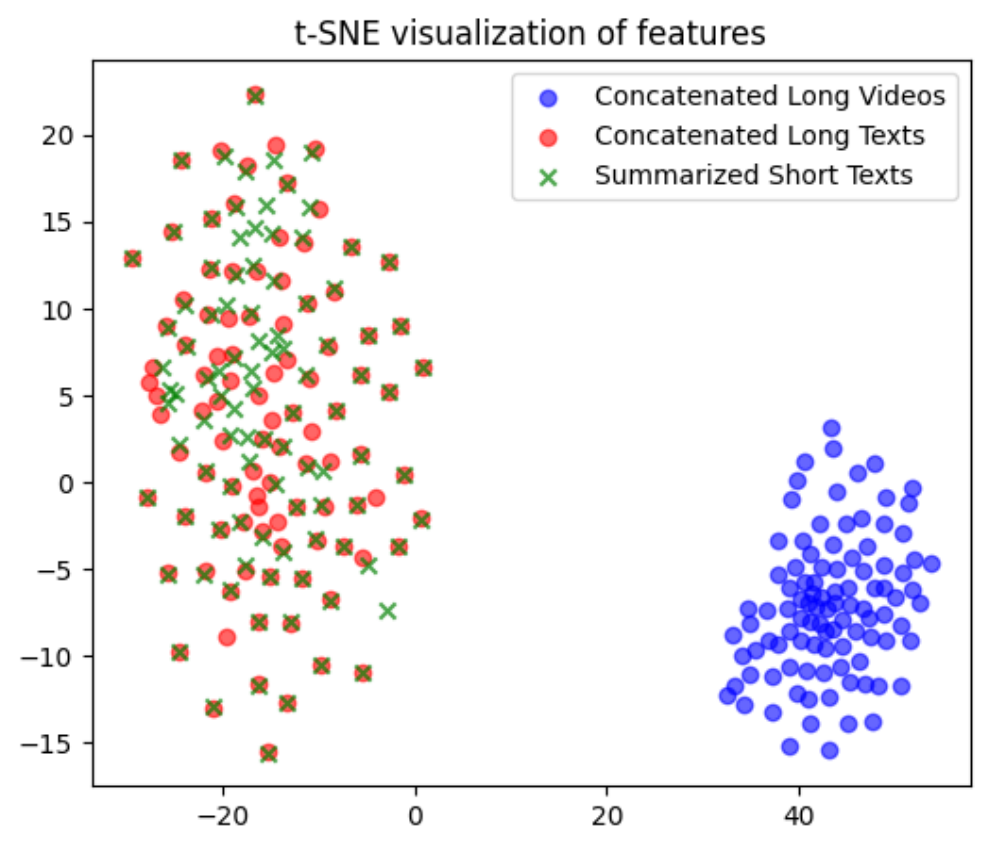}
    \caption{t-SNE visualization of the CLIP-based features for the sampled 100 concatenated long videos, concatenated long texts, and summarized short texts.}
    \label{supple-fig-sim}
\end{figure}

We randomly sample 100 concatenated long videos along with their corresponding long texts and summarized short texts, and compute their CLIP-based~\cite{clip} features. Next, we calculate the average cosine similarities between the full set of 1M text features and the 100 sampled video-text feature pairs, as shown in Table~\ref{supple-sim}. The t-SNE visualization of the 100 sampled features is also provided in Figure~\ref{supple-fig-sim}, where we observe the clustering of the features.

From the similarities in Table~\ref{supple-sim}, we observe that the long-video-long-text pairs (LVLT) and long-video-short-text pairs (LVST) exhibit similar levels of similarity in their CLIP-based features (0.220$\pm$0.037 vs. 0.215$\pm$0.033, respectively). The relatively low similarity between video and text features can be attributed to the domain gap between the image-pretraining dataset of the CLIP model and this video dataset. However, both the t-SNE plot and the cosine similarity scores show a strong resemblance between the summarized short texts and the original concatenated long texts, reflected in the much higher similarity score of 0.791. These findings suggest that the summarized short texts preserve enough semantic information from the long texts to effectively serve as positive samples with the long videos for subsequent alignment learning.

\begin{table}[t]
\caption{The average cosine similarities of CLIP-based features between concatenated long videos, concatenated long texts, and summarized short texts in the pretrained dataset.}
\scalebox{0.74}{
\begin{tabular}{cccc}
\hlineB{3}
  Features  & \begin{tabular}[c]{@{}c@{}}concat. long videos \\ concat. long texts\\ (100 samples)\end{tabular} & \begin{tabular}[c]{@{}c@{}}concat. 
 long videos \\ sum. short texts\\ (100 samples)\end{tabular} & \begin{tabular}[c]{@{}c@{}}concat. long texts\\ sum. short texts\\ (1M)\end{tabular} \\ \hline
\begin{tabular}[c]{@{}c@{}}Avg$\pm$Std\\ Cos Sim\end{tabular} & 0.220$\pm$0.037                                                                     & 0.215$\pm$0.033                                                                      & 0.791$\pm$0.084                                                            \\ \hlineB{3}
\end{tabular}
\label{supple-sim}
}
\end{table}

\begin{figure}
    \centering
    \includegraphics[width=\columnwidth]{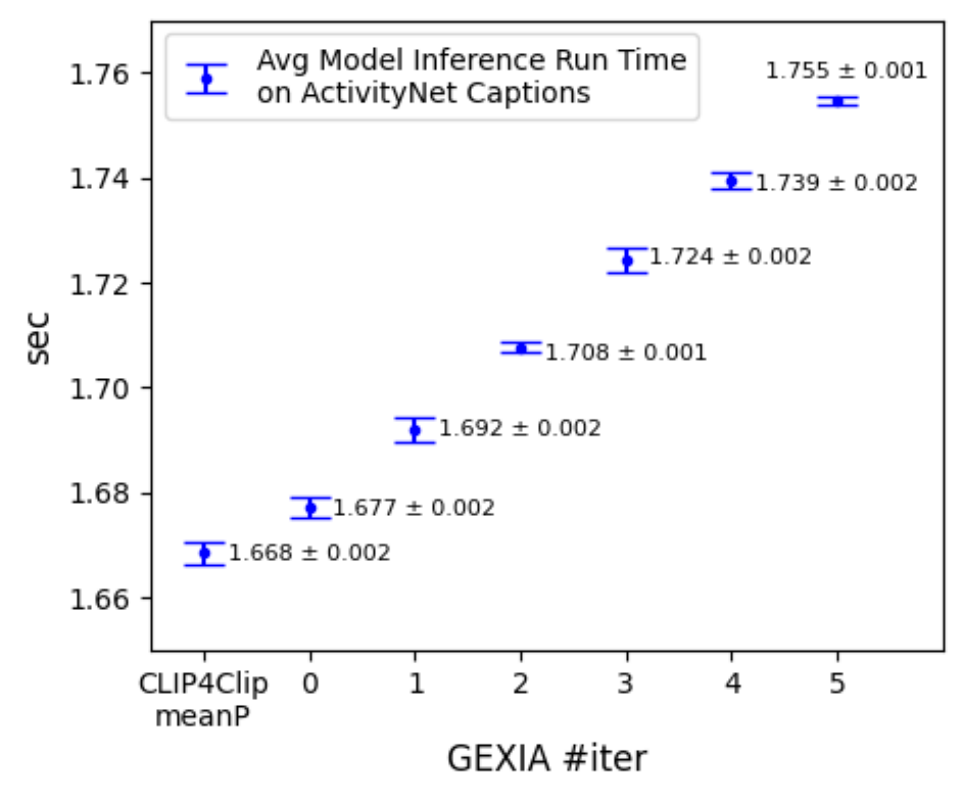}
    \caption{Average model inference run time on ActivityNet Captions across different $\#iter$ setups for GEXIA, compared to CLIP4Clip (mean pooling setup).}
    \label{supple-run-time}
\end{figure}

\section{Retrieval Complexity and Inference Efficiency}
Our GEXIA method incorporates separate video and text branches, similar in structure to CLIP4Clip~\cite{clip4clip}, maintaining the same efficient retrieval complexity of $\mathcal{O}(N_vN_t)$, where $N_v$ represents the number of candidate videos and $N_t$ is the size of the text query set. The video and text IAMs are connected with the video and text branches respectively, thus the $\#iter$ setups do not affect the retrieval complexity. In contrast, retrieval-specific methods like X-CLIP~\cite{xclip} introduce cross-modal and fine-grained features before retrieval, resulting in a much higher complexity of $\mathcal{O}(N_vN_tN_fN_w)$, where $N_f$ denotes the number of frames per video, and $N_w$ represents the average length of words in the texts. Additionally, many retrieval-specific models~\cite{frozen, xclip, promptswitch, ucofia, centerclip} are finetuned exclusively on retrieval datasets, which limits their applicability as foundational models for broader video understanding and classification tasks.

To further demonstrate the efficiency of our model, we present the average inference run time across different $\#iter$ setups, compared to CLIP4Clip (mean pooling setup)~\cite{clip4clip}. The experiment was conducted on one NVIDIA Tesla-V100 GPU with a batch size of 64 using the ActivityNet Captions dataset. As shown in Figure~\ref{supple-run-time}, our model's run time is only slightly longer than CLIP4Clip, due to the smaller feature size and minimal computational overhead introduced by the iterative approximation process. Specifically, with $\#iter$=3, our model achieves an 11.8\% relative improvement in text-to-video R@1 on the ActivityNet Captions dataset (according to Table 1 in the paper), while requiring only 3.3\% more inference time compared to CLIP4Clip. These results highlight the efficiency of our model, offering substantial performance gains with only a modest increase in computational cost.

\section{Comparison of Computational Costs}
We compare the computational costs between our model and various baseline models, using FLOPs (Floating Point Operations) per frame as the metric. As illustrated in Table~\ref{gflops}, our model demonstrates one of the lowest FLOPs among the compared models, largely due to the use of a compact backbone (ViT-B/32) for the local video encoder. Notably, the ViCLIP model incurs a computational cost that is 17$\times$ higher than our model. This lower computational cost makes our model not only efficient in terms of inference time but also in terms of overall computational resource consumption, enhancing its applicability in real-world scenarios.

\section{Detailed Dataset Statistics and $\#iter$ Setups}
We provide the detailed average video/text lengths of each downstream dataset and the corresponding $\#iter$ setups in Table~\ref{tab:set}. We set these $\#iter$s based on the average length of the videos and texts of the given datasets, where $\#iter=3$ for long video/text data and $\#iter=1$ for short ones.


\begin{table}[t]
\centering
\footnotesize
\caption{GFLOPs per frame of our model and other baselines.}
\begin{tabular}{l | c | c }
        \hlineB{3}
        Model  & Vision Backbone & GFLOPs  \\\hline
        CLIPBert~\cite{clipbert} & $\text{Res}_{50}$ & 4.1 \\
        TACo~\cite{taco} & $\text{Res}_{152}$ & 11.6 \\
        Frozen~\cite{frozen} & TS4mer & 44.5 \\
        LF-VILA~\cite{lfvila} & $\text{Swin}_B$ & 9.3 \\
        HiTeA~\cite{hitea} & $\text{MViT}_B$ & 8.2 \\
        LocVTP~\cite{locvtp} & $\text{ViT}_{B16}$ & 17.6 \\
        ViCLIP~\cite{internvid} & $\text{ViT}_{L14}$ & 81.1 \\
        VideoPrism\cite{videoprism} & $\text{ViViT}_{B}$ &  $\geq$ 142.0~\cite{vivit}\\
        Ours & $\text{ViT}_{B32}$ & 4.7 \\
        \hlineB{3}
    \end{tabular}

\label{gflops}
\end{table}

\begin{table}[t]
    \footnotesize
    \centering
    \caption{Average video/text lengths of the seven downstream datasets and the corresponding $\#iter$ setups.}
    \scalebox{1}{
    \begin{tabular}{lccc}
        \hlineB{3}
        Dataset    & Video (sec) & Text (\#word) & $\#iter$ (V-T) \\ \hline
        ActivityNet &  180  & 49.2 & 3-3    \\
        MSR-VTT & 14.8 & 9.3   & 1-1   \\
        LSMDC & 4.7 & 9.7  & 1-1  \\
        LVU  & 120 & N/A & 3 (only V) \\
        COIN & 141.6 & N/A & 3 (only V) \\
        Charades-Ego & 31.5 & 3.9  & 3-1  \\
        How2QA & 17.5&  16.0  & 1-1 \\ \hlineB{3}
        \end{tabular}}
    \label{tab:set}
\end{table}

\section{Extension to a New Granularity: Image-Text Data}
\begin{table}[t]
\centering
\footnotesize
\caption{Zero-shot ImageNet~\cite{imagenet} classification accuracy results. All models utilize a ViT-B/32 backbone. ZS: Zero-shot; IT: Image-Text data pairs; VT: Video-Text data pairs; $\#iter$: Video-Text iteration number.}
\scalebox{0.9}{
\begin{tabular}{lccc}
\hlineB{3}
Method                      & Pretraining Dataset\ \                      &\#PT Granularities\ \         & ZS Acc. \\ \hlineB{1}
CLIP~\cite{clip}            & YFCC-15M~\cite{yfcc}           & 1 (IT only)                     & 32.8 \\
SLIP~\cite{slip}            & YFCC-15M~\cite{yfcc}           & 1 (IT only)     & 34.3        \\
FILIP~\cite{filip}          & YFCC-15M~\cite{yfcc}           & 1 (IT only)     & 39.5        \\

Ours ($\#iter$: 1-1)        & InternVid-10M~\cite{internvid}        & 3 (VT only)      & 32.5        \\
Ours ($\#iter$: 3-1)        & InternVid-10M~\cite{internvid}        & 3 (VT only)      & 31.0        \\
Ours ($\#iter$: 1-1)               & InternVid-10M~\cite{internvid} & 4 (VT+IT)      & 33.9        \\
Ours ($\#iter$: 3-1)               & InternVid-10M~\cite{internvid} & 4 (VT+IT)      & 31.5        \\
Ours ($\#iter$: 0-1)               & InternVid-10M~\cite{internvid} & 4 (VT+IT)      & \textbf{40.6}        \\ \hlineB{3}
\end{tabular}}
\label{supple_image}
\end{table}

To further explore the generalization capability of our method, we extended its application to include a new type of data granularity: images to short texts, where images can be seen as one-frame videos. This expansion involves generating image-text pairs for pretraining, achieved by the Video Compression ($\Theta_v$). Here we extract the middle frame from each short video in the InternVid-10M-FLT~\cite{internvid} dataset and pair it with the corresponding text of the video. As such, we compose 10M image-text pairs. Following this, we take the model that was initially pretrained on video-text pairs and proceed with further pretraining using these newly formed image-text pairs for 5 epochs. Additionally, we set $\#iter=0$ for the video branch to differentiate the granularities of the image and video and keep the text $\#iter$ as 1. 

\begin{figure*}[t]
     \centering
     \includegraphics[width=.83\textwidth]{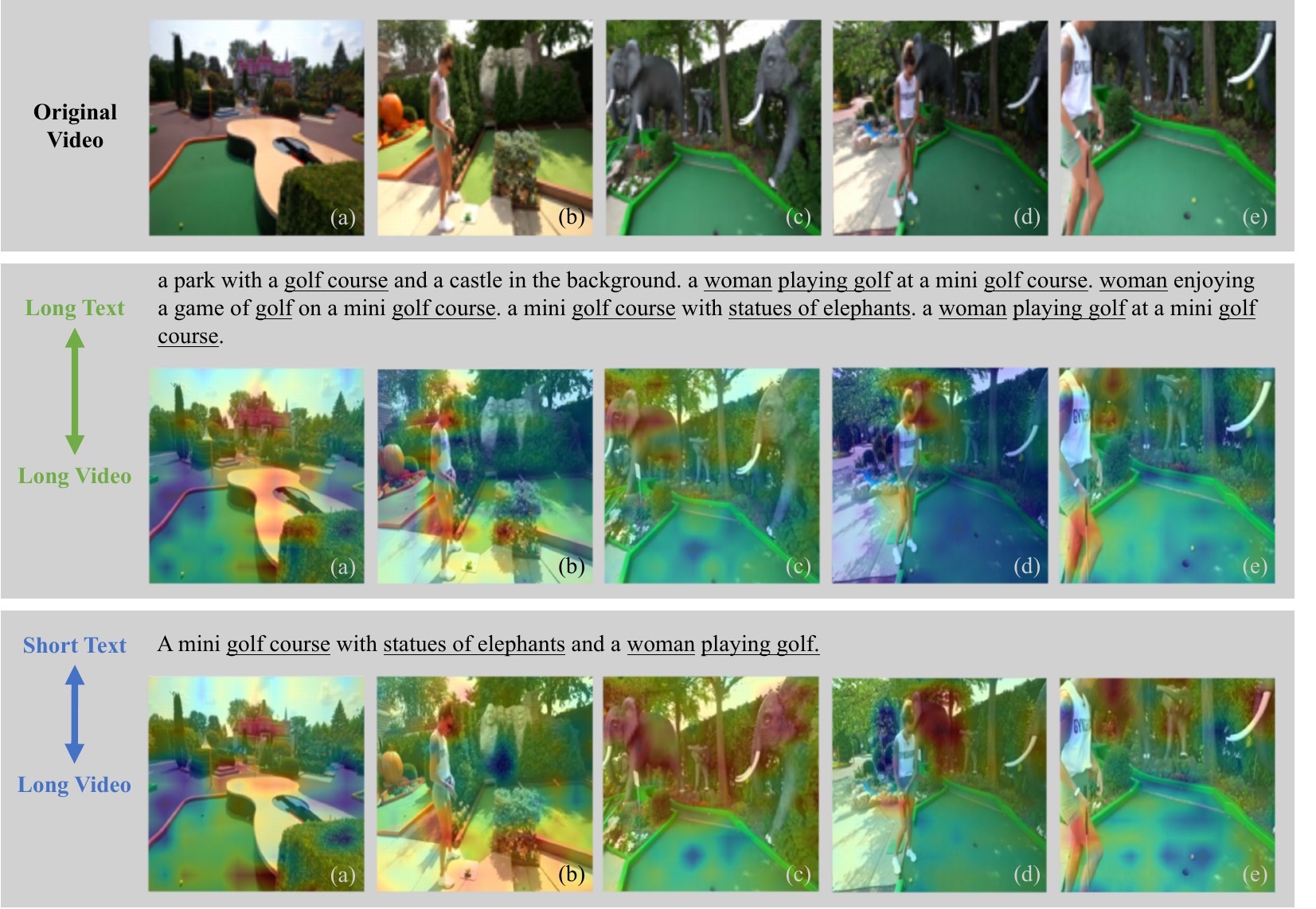}
     \caption{The visualization of alignment scores for long-video-long-text and long-video-short-text pairs. Given the same video, our GEXIA method is able to capture and align different information according to the input texts. Areas highlighted in {\color{red} red} indicate regions of higher alignment scores, whereas the {\color{blue} blue} regions represent areas with lower alignment scores.}
     \label{fig4}
\end{figure*}

\begin{figure*}[!t]
     \centering
     \includegraphics[width=.84\textwidth]{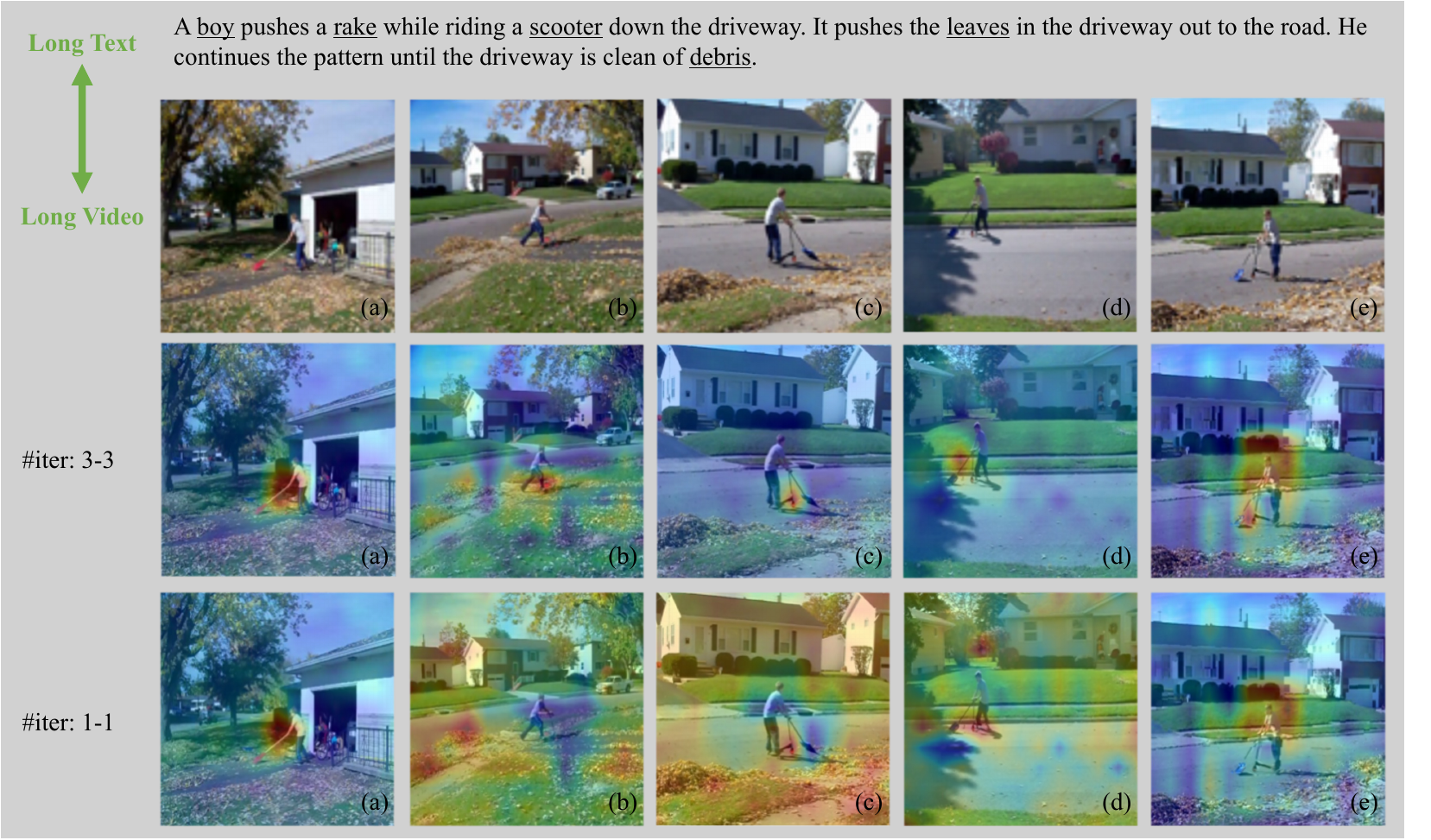}
     \caption{The visualization of alignment scores for different $\# iter$ settings. The 3-3 setup of $\#iter$ works better than the 1-1 in this long-video-long-text case. Areas highlighted in {\color{red} red} indicate regions of higher alignment scores, whereas the {\color{blue} blue} regions represent areas with lower alignment scores.}
     \label{fig5}
\end{figure*}

After completion of the pre-training phase, we use the zero-shot ImageNet~\cite{imagenet} classification task as a benchmark to assess the generalizability of our GEXIA method. For this purpose, to implement the pretrained model for zero-shot image classification, following the CLIP setup~\cite{clip}, we employ a prompt template: ``A video of a \{label\}.'' to transform the problem into zero-shot image-text retrieval. The results of the experiment are shown in Table~\ref{supple_image}. Our model demonstrates remarkable generalization capability to the new granularity of image-short-text, surpassing reference models targeting image tasks by a large margin. The comparison across our five models shows that setting $\#iter=0$ for image inputs and incorporating image-text pairs into the pretraining data leads to the highest performance for image-based tasks. This finding further affirms that assigning appropriate iteration numbers based on the granularity of the input, along with pretraining data of the targeted granularity, is the key to achieving effective multi-grained visual-language alignments.

\section{Qualitative Study of Cross-modal Alignments}
\label{sec:visualization}
We qualitatively study the video-text alignments of our GEXIA models by visualizing pixel-level alignment scores across temporal and spatial dimensions. Given an input video $v$ and text $t$, we start by computing the similarity value $S$ between the output video and text embeddings. Next, we create a modified version of the video by masking a small patch at coordinates $[h,w]$ in one frame $t$ of the video, resulting in $v^{\text{mask}}_{t,h,w}$. We then calculate the similarity value $S^{\text{mask}}_{t,h,w}$ between the embeddings of the masked video and the text. 

The difference between $S_{t,h,w}$ and $S^{\text{mask}}_{t,h,w}$ is defined as the alignment score at the patch level, indicating the reduction in alignment when the mask is applied. We apply a $32\times32$ mask patch across the video frames in a sliding window fashion with a $16$-pixel stride, producing a $13\times13$ matrix of patch-level alignment scores for each frame. We then resize these scores to match the original input video dimensions, resulting in the pixel-level alignment scores. We visualize the pixel-level scores in Figure~\ref{fig4} and Figure~\ref{fig5}.


\subsection{Alignments with Long and Short Texts}

In Figure~\ref{fig4}, when presented with the same long video, we observe notable differences between long text and short text inputs. For the long text, which contains more detailed information, the alignment scores are dispersed more uniformly across various regions and objects within the video. This indicates a comprehensive integration of video content with extensive textual details. Conversely, in the case of short text, the alignment scores are more focused on specific key elements mentioned in the text. For example, the model concentrates on "the status of elephants" in frames (c) and (e), and on "a woman playing golf" in frame (b). This pattern reveals that our model is adept at aligning video content with both long- and short-text inputs, effectively adjusting its focus based on the granularity of the text. 
 
\subsection{Alignments with Different Iteration Numbers}

In Figure~\ref{fig5}, we further study the impact of the iteration number $\#iter$ in a qualitative way. This assessment involves a comparison of visualized pixel-level alignment scores using two $\#iter$ settings for a given pair of long video and long text. The figure reveals that when $\#iter$ is set to 1-1, the model struggles to identify key details in the text, notably missing elements like "scooter" and "rake" in frames (b), (c), and (d). Additionally, the alignment scores appear more randomly scattered across the frames, suggesting suboptimal alignment in this configuration. On the other hand, the $\#iter$ set to 3-3 shows a contrast. This setup enables the model to detect all critical details in the text, reflected in high alignment scores for corresponding objects in the video. This difference in performance between the two $\#iter$ settings not only highlights the significant role of iteration numbers during inference but also reaffirms the adaptability of our model to multi-grained video-text pairs.


%
%
%

{\small
\bibliographystyle{ieee_fullname}
\bibliography{egbib}
}

\end{document}